\documentclass[lettersize,journal]{IEEEtran}
\usepackage{amsmath,amsfonts}
\usepackage{algorithmic}
\usepackage{algorithm}
\usepackage{array}
\usepackage[caption=false,font=normalsize,labelfont=sf,textfont=sf]{subfig}
\usepackage{textcomp}
\usepackage{stfloats}
\usepackage{url}
\usepackage{verbatim}
\usepackage{graphicx}
\usepackage{cite}
\usepackage[hidelinks]{hyperref}
\usepackage[capitalise,nameinlink]{cleveref}

\usepackage{booktabs}
\usepackage{multirow}
\usepackage{array}
\usepackage{makecell}
\usepackage{threeparttable}
\usepackage[table]{xcolor}

\usepackage{capt-of}

\begin{document}

\title{GroupForward: Building Referable 3D Scenes via Instance-Grouped Feed-Forward Gaussian Splatting}

\author{
Qijian Tian,
Zimeng Wu,
Xuhong Wang,
Lizhuang Ma,
and Xin Tan
\thanks{Qijian Tian and Lizhuang Ma are with the School of Computer Science, Shanghai Jiao Tong University, Shanghai 200240, China.}
\thanks{Zimeng Wu is with the School of Computer Science and Engineering, Beihang University, Beijing 100191, China.}
\thanks{Xuhong Wang and Xin Tan are with Shanghai Artificial Intelligence Laboratory, Shanghai 200232, China.}
\thanks{Xin Tan is also with the School of Computer Science and Technology, East China Normal University, Shanghai 200241, China.}
}




\maketitle

\begin{abstract}
Simultaneously reconstructing and understanding 3D environments is essential for embodied agents. Toward this goal, feed-forward semantic 3D Gaussian Splatting (3DGS) efficiently constructs semantic scene representations from sparse multi-view observations. However, existing methods lack explicit instance discrimination and mainly support category- or phrase-based semantic queries. To this end, we propose GroupForward, an instance-grouped feed-forward Gaussian splatting model that reconstructs geometry, appearance, instance structure, and semantics from sparse, unposed, and uncalibrated multi-view images. Unlike existing methods that attach high-dimensional semantic features to each Gaussian, GroupForward learns compact instance embeddings that group Gaussians into cross-view consistent 3D instances, reformulating feed-forward semantic 3DGS from per-Gaussian semantic feature rendering to instance-level semantic aggregation and propagation. Building on these instance groups, we further propose a Referential Scene Reasoning Framework (RSRF) for complex 3D referring segmentation. RSRF constructs an instance-grouped 3D scene graph and retrieves candidate instances for a given referring expression. A vision-language model then reasons over structured instance evidence and multi-view observations to identify the referred instance among the candidates. RSRF thereby extends language interaction from simple semantic querying to complex referential scene reasoning. Experiments on semantic reconstruction and referential reasoning demonstrate the effectiveness of our instance-grouped reconstruction and reasoning framework. \par
\end{abstract}

\begin{IEEEkeywords}
Semantic 3D Reconstruction, 3D Gaussian Splatting, Scene Understanding, Referential Reasoning.
\end{IEEEkeywords}
\section{Introduction}

\begin{figure*}[ht]
    \centering
    \includegraphics[width=0.98\textwidth]{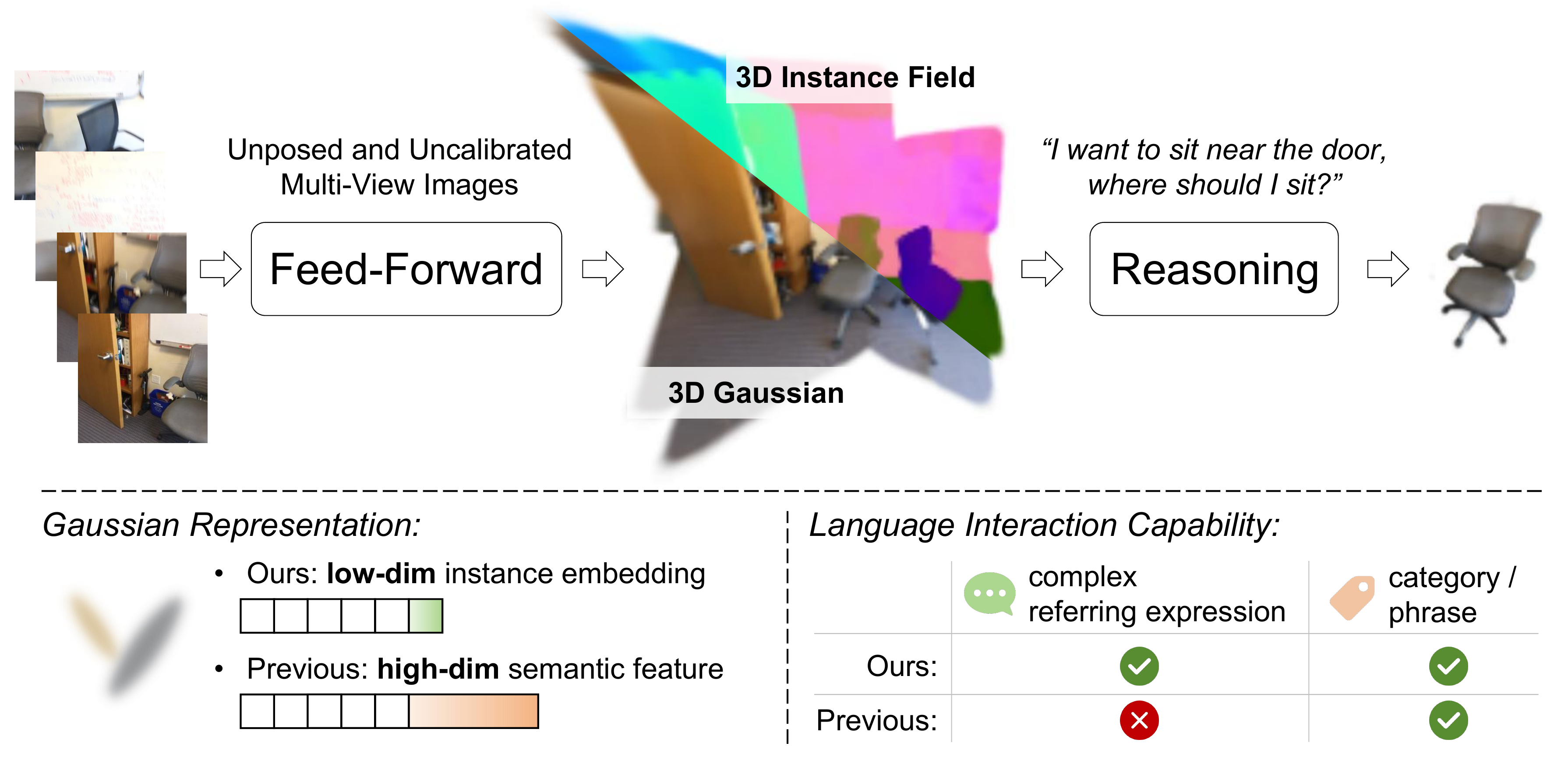}
    \caption{The proposed instance-grouped formulation for feed-forward semantic 3D Gaussian Splatting builds referable scenes with low-dimensional instance embeddings rather than attaching high-dimensional semantic features, enabling complex referential reasoning beyond category- or phrase-level queries.}
    \label{fig:teaser}
\end{figure*}

Embodied intelligence requires agents to perceive, understand, and execute tasks in real-world 3D environments. To support these capabilities, agents need to rapidly construct the surrounding scene from limited observations and reason about spatial relationships. Scene reconstruction can recover 3D scenes from 2D observations, providing the foundation for such spatial perception and understanding. However, geometry and appearance alone are insufficient for semantic understanding and decision-making. Therefore, 3D scene representations for downstream tasks should further incorporate semantic information and spatial relationships, extending reconstructed scenes from renderable 3D models to structured representations that support scene understanding, referential reasoning, and embodied interaction. \par

To achieve this goal, semantic 3D reconstruction incorporates semantic features into 3D representations, enabling semantic understanding of reconstructed scenes. Recently, 3D Gaussian Splatting~\cite{3dgs} (3DGS) has been widely adopted due to its efficient rendering, explicit representation, and flexibility in attaching semantic attributes. However, most existing semantic 3DGS methods~\cite{langsplat, feature3dgs, legaussians, opengaussian, langsplatv2} rely on per-scene optimization, where geometry, appearance, and semantic attributes are optimized for each scene. Such a process is time-consuming and typically requires dense observations, limiting their applicability to sparse-view reconstruction and real-time interaction. This motivates feed-forward semantic 3DGS reconstruction~\cite{lsm, uni3r}, which directly predicts Gaussian scene representations of geometry, appearance, and semantics from sparse input views. \par

Nevertheless, existing feed-forward semantic 3DGS reconstruction methods exhibit two major limitations. First, most methods formulate semantic representation as per-Gaussian feature embeddings, where high-dimensional language-aligned features are attached to individual Gaussians and rendered into 2D semantic maps. Although such representations associate Gaussians with semantic features or open-vocabulary concepts, they do not explicitly group Gaussians into object instances, making it difficult to distinguish multiple instances within the same category and preserve cross-view identity consistency. Second, existing methods mainly support category-level or phrase-based semantic queries, such as object category names or simple phrases, through semantic similarity matching. However, real-world language instructions often involve complex referring expressions, such as “the trash bin near the cabinet” or “the place in the scene where one can wash hands,” which require joint reasoning over object identities, spatial relationships, functional attributes, and contextual information. These limitations make it difficult for existing methods to support complex language-based referential reasoning. \par

To address these limitations, we introduce an instance-grouped scene representation into feed-forward 3DGS, where instances serve as structured entities that bridge Gaussian primitives and language reasoning. As illustrated in Fig.~\ref{fig:teaser}, our framework reconstructs cross-view consistent 3D instance groups from unposed and uncalibrated multi-views, discriminating instances within the same category. These instance groups further provide a structured scene representation for complex referential reasoning, grounding referring expressions to explicit 3D instance entities using geometric, appearance, semantic, and spatial cues. \par

Specifically, we first propose GroupForward, an instance-grouped feed-forward Gaussian splatting model. In contrast to existing methods that directly embed high-dimensional semantic features into each Gaussian, GroupForward learns low-dimensional instance embeddings with rendering-consistent instance supervision, grouping Gaussian primitives into cross-view consistent 3D instances. Semantic information is then aggregated and propagated at the instance level. We extract semantic features from context views using a 2D open-vocabulary model and aggregate them into a semantic prototype for each 3D instance. In the target view, rendered instance partitions are combined with these prototypes to produce view-consistent open-vocabulary semantic predictions. GroupForward thus reformulates semantic 3D reconstruction from high-dimensional per-Gaussian feature rendering to instance-level semantic aggregation and propagation. \par

Moreover, we build a Referential Scene Reasoning Framework (RSRF) on the 3D instance groups produced by GroupForward to understand complex natural-language referential expressions and achieve 3D referring segmentation. The framework constructs an instance-grouped scene graph, where each node represents a cross-view consistent 3D instance with its semantic prototype, multi-view masks, 3D position, and visibility, while edges encode spatial and geometric relations between instances. Given a referential expression, RSRF retrieves candidate instances using semantic, geometric, and graph cues and organizes the corresponding information into an evidence bundle. A vision-language model (VLM) then reasons over existing instance nodes to select the target, rather than generating new segmentation masks. The target instance and the corresponding cross-view consistent masks are directly returned. By grounding language understanding in a stable 3D instance graph, RSRF extends feed-forward semantic 3DGS from category-level semantic querying to instance-grounded scene reasoning and 3D referring segmentation. \par

Our main contributions are summarized as follows:

\begin{itemize}
\item We propose Instance-Grouped Feed-Forward Gaussian Splatting (GroupForward), which shifts feed-forward semantic 3DGS from per-Gaussian high-dimensional semantic feature rendering to instance-grouped reconstruction based on low-dimensional instance embeddings, enabling explicit instance-level scene structures from sparse input views. \par

\item Building on the 3D instance groups, we propose a Referential Scene Reasoning Framework (RSRF) based on a 3D instance scene graph, extending semantic 3DGS from category-level or phrase-based semantic queries to complex natural-language referential reasoning and 3D referring segmentation. \par

\item Experiments on sparse-view reconstruction, semantic reconstruction, and referential reasoning show that our method outperforms existing methods and demonstrate the effectiveness of instance-grouped feed-forward reconstruction and 3D instance-based reasoning. \par

\end{itemize}

\section{Related Work}

\subsection{Feed-Forward 3D Reconstruction}

Typical 3D reconstruction paradigms, such as Structure-from-Motion~\cite{sfm}, Neural Radiance Fields~\cite{nerf} (NeRF), and 3D Gaussian Splatting~\cite{3dgs} (3DGS), rely on per-scene optimization to recover scene geometry and appearance. Although these methods achieve high reconstruction quality, they often require dense observations and take minutes or even hours to reconstruct a single scene, making them less suitable for embodied intelligence~\cite{conceptgraphs,3dmem, mtu3d}, where 3D representations need to be rapidly constructed from sparse observations for perception, planning, and interaction. This has motivated feed-forward 3D reconstruction~\cite{pixelnerf, dust3r, mast3r, pixelsplat, mvsplat, drivingforward, vggt, anysplat}, which learns generalizable cross-scene priors and directly infers 3D representations from sparse input views in a single forward pass without costly per-scene optimization. Existing methods use different 3D representations. Pointmap-based approaches mainly recover geometry from unposed sparse views~\cite{dust3r, mast3r, vggt}, while NeRF- and 3DGS-based approaches additionally model appearance for novel-view synthesis~\cite{pixelnerf, pixelsplat, mvsplat, anysplat}. However, they generally do not incorporate semantics into reconstructed scenes, limiting downstream scene understanding and reasoning. \par

\subsection{Semantic 3D Reconstruction}

Semantic 3D reconstruction extends typical 3D reconstruction by incorporating semantic information into 3D scene representations. Beyond geometry and appearance, it associates 3D representations such as points, radiance fields, or 3D Gaussians with semantic labels or language-aligned features, enabling semantic understanding, open-vocabulary querying, or language-guided interaction. Similar to 3D reconstruction, semantic 3D reconstruction methods initially rely on per-scene optimization. NeRF-based methods~\cite{semanticnerf, dff} and 3DGS-based methods~\cite{langsplat, feature3dgs, legaussians, opengaussian, langsplatv2} attach semantic or language-aligned features to 3D representations for semantic rendering and querying. Recent studies have further explored feed-forward semantic 3D reconstruction to improve efficiency and practicality. Some methods~\cite{iggt, ff3r, unite, fast3dis} incorporate semantics into pointmap-based feed-forward reconstruction models. However, point-based representations have limited appearance modeling capability for high-quality rendering and novel-view synthesis. Other methods~\cite{lsm, semanticsplat, spatialsplat, gsemsplat, uniforward, uni3r, fleg} perform feed-forward semantic reconstruction with 3D Gaussians, jointly reconstructing geometry, appearance, and semantics. Most of these methods attach high-dimensional semantic features to each Gaussian and distill rendered feature maps from 2D semantic features. This design introduces semantics into 3DGS but lacks explicit instance modeling, limiting the ability to distinguish same-category instances. Rendering high-dimensional semantic features also increases computational cost and may lead to feature degradation. SIU3R~\cite{siu3r} avoids dense feature rendering through query-based mask prediction, but its reliance on closed-set ground-truth semantic labels limits open-vocabulary capability. In contrast, GroupForward replaces high-dimensional per-Gaussian semantic features with low-dimensional instance embeddings and aggregates open-vocabulary semantic prototypes at the instance level. This enables view-consistent open-vocabulary semantic prediction without ground-truth semantic labels while providing 3D instance entities for downstream referential reasoning. \par

\subsection{3D Scene Understanding and Reasoning}

Open-world scene understanding has been widely studied in 2D vision through CLIP-based methods~\cite{clip, lseg} and has recently been extended to 3D. SceneSplat~\cite{scenesplat} and SceneSplat++~\cite{scenesplat++} introduce language-aligned 3D Gaussian representations for open-world 3D scene understanding. However, these methods mainly support category-level or phrase-based semantic queries and struggle with complex natural-language instructions. Existing semantic 3D reconstruction methods~\cite{dff, feature3dgs, lsm, uni3r} also have this limitation, as they primarily focus on semantic querying rather than complex referential reasoning. MLLMs provide stronger language understanding and reasoning capabilities~\cite{gpt4o, qwen3vl, sa2va, efficientmllm}, but lack explicit 3D perception and thus fail to directly ground language in 3D representations. ReferSplat~\cite{refersplat} explores referring 3D segmentation on 3DGS, while REALM~\cite{realm} further leverages MLLMs for this task. However, they both rely on dense observations and per-scene optimization. In contrast, our RSRF combines feed-forward semantic reconstruction with MLLM-based reasoning to enable efficient referential scene understanding and reasoning from sparse observations. \par
\section{Method}

\begin{figure*}[ht]
    \centering
    \includegraphics[width=0.98\textwidth]{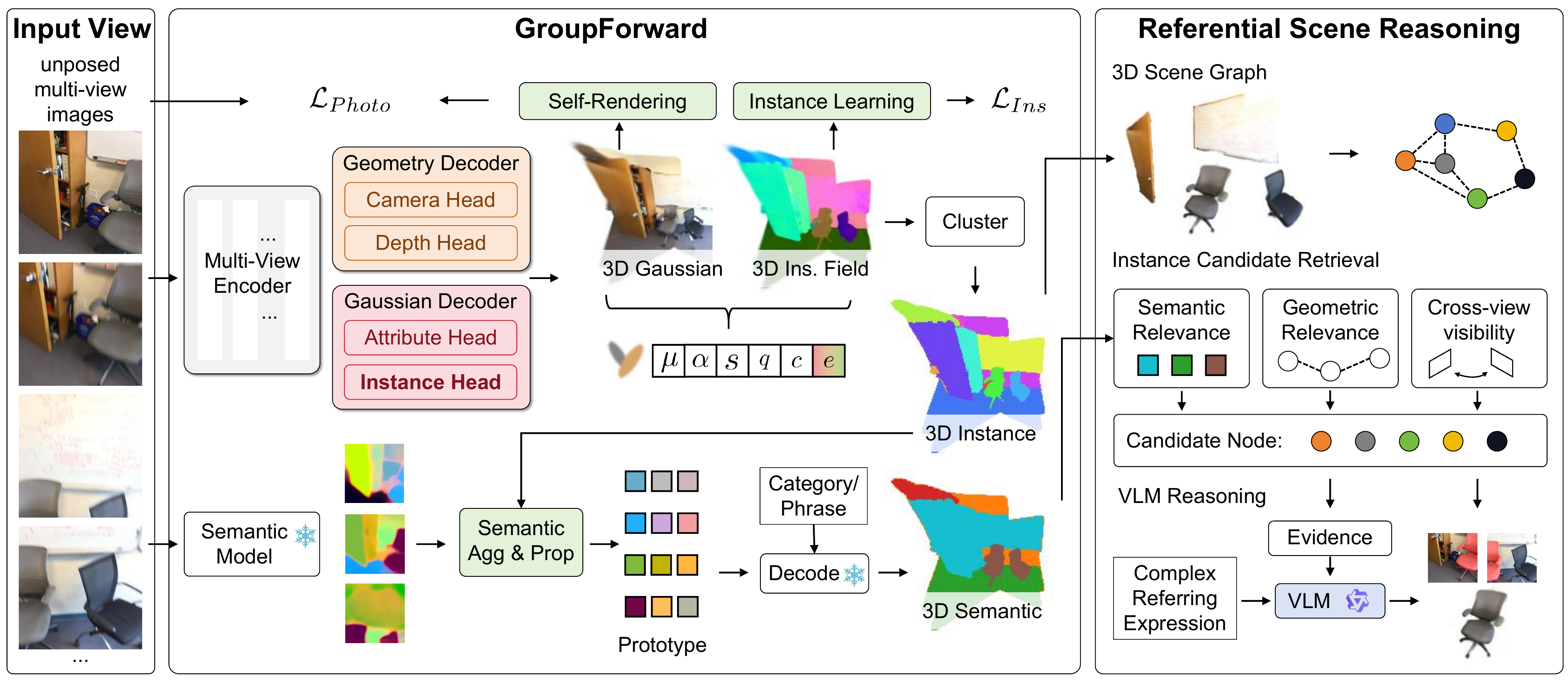}
    \caption{Overview of GroupForward and the Referential Scene Reasoning Framework (RSRF). Given sparse, unposed, and uncalibrated multi-view images, GroupForward reconstructs cross-view consistent 3D instance groups with compact instance embeddings and aggregates open-vocabulary semantics at the instance level. RSRF further organizes the instance groups into a 3D scene graph and integrates structured graph evidence with multi-view observations for VLM-based complex referential reasoning and 3D referring segmentation.}
    \label{fig:main}
\end{figure*}

\subsection{Overview}
As illustrated in Fig.~\ref{fig:main}, our method consists of GroupForward (Sec.~\ref{sec:GroupForward}) for instance-grouped feed-forward 3D reconstruction and the Referential Scene Reasoning Framework (RSRF; Sec.~\ref{sec:rsrf}) for referential scene reasoning. The key insight is to use persistent 3D instance groups as intermediate entities to bridge low-level Gaussian reconstruction and high-level language reasoning. GroupForward jointly predicts cameras and Gaussian primitives with compact instance embeddings, which are then clustered into cross-view consistent 3D instance groups. Open-vocabulary features from the context views are aggregated within each group to form an instance-level semantic prototype. Built on this instance-grouped representation, RSRF organizes the instance groups into a 3D scene graph, retrieves candidate instances based on the input expression, and employs a VLM to identify target 3D entities using structured graph evidence and multi-view observations. The selected instances are then directly mapped to their cross-view masks without additional segmentation. \par

\subsection{Instance-Grouped Feed-Forward Gaussian Splatting}
\label{sec:GroupForward}

GroupForward predicts Gaussian primitives with compact instance embeddings in a feed-forward manner. Its design further includes gauge-aligned self-rendering, rendering-consistent instance learning, instance-level semantic aggregation and propagation, and staged curriculum training. \par

\textbf{Feed-Forward Gaussian Splatting with Instance Embeddings.}
The feed-forward model uses a shared multi-view encoder followed by two decoders with complementary roles. The geometry decoder comprises camera and depth heads for geometry prediction. The Gaussian decoder also comprises two lightweight heads. The attribute head decodes pixel-aligned attributes for standard Gaussians, while the instance head predicts compact embeddings for instance grouping. Specifically, each input pixel is lifted to a Gaussian primitive,
\begin{equation}
    {G}_{i,p} =
    \left(
    \boldsymbol{\mu}_{i,p},
    \alpha_{i,p},
    \mathbf{s}_{i,p},
    \mathbf{q}_{i,p},
    \mathbf{c}_{i,p},
    \mathbf{e}_{i,p}
    \right),
\end{equation}
where $\boldsymbol{\mu}_{i,p}$, $\alpha_{i,p}$, $\mathbf{s}_{i,p}$, $\mathbf{q}_{i,p}$, and $\mathbf{c}_{i,p}$ respectively denote the 3D center, opacity, scale, rotation, and spherical-harmonic appearance coefficients of the Gaussian associated with pixel $p$ in view $i$. In contrast to standard 3D Gaussian primitives, we augment each primitive with an additional instance embedding $\mathbf{e}_{i,p}\in\mathbb{R}^{d_e}$. The embedding dimension $d_e$ is set to be substantially lower than typical semantic feature dimensions, since $\mathbf{e}_{i,p}$ is intended to encode instance grouping cues rather than semantics. The Gaussian decoder integrates multi-scale image features and predicted geometric cues to recover scene geometry and appearance, and the instance head combines multi-scale backbone features with predicted depth to promote embeddings aligned with object boundaries and 3D structure. \par

\textbf{Gauge-Aligned Self-Rendering.}
Feed-forward Gaussian reconstruction is commonly trained with photometric supervision between observed and synthetic images rendered with ground-truth camera parameters. While providing a reliable reconstruction signal, this supervision does not account for pose estimation errors and therefore does not explicitly enforce consistency between the reconstructed scene and the predicted camera poses. To address this limitation, we introduce gauge-aligned self-rendering, which supervises the reconstructed scene under self-predicted camera poses. Specifically, we construct the Gaussian scene from the context views and perform an additional stop-gradient pose inference pass over both the context and target views, producing the joint pose predictions. Since the context-only and joint pose predictions may differ in global scale, we align the latter to the former using their shared context views. Let $\mathbf{t}^{c}_{i}$ and $\mathbf{t}^{r}_{i}$ denote the translation vectors predicted for context view $i$ by the context-only reconstruction pass and the joint pose inference pass, respectively. The alignment scale is computed as
\begin{equation}
    \gamma =
    \frac{
        \sum_{i \in \mathcal{C}}
        \left\lVert \mathbf{t}^{c}_{i} \right\rVert_{2}
    }{
        \sum_{i \in \mathcal{C}}
        \left\lVert \mathbf{t}^{r}_{i} \right\rVert_{2}
    },
\end{equation}
where $\mathcal{C}$ denotes the context-view set. We scale the translations of all jointly predicted poses as $\widetilde{\mathbf{t}}^{r}_{v}=\gamma\mathbf{t}^{r}_{v}$, where $v\in\mathcal{C}\cup\mathcal{T}$, and render the context-built Gaussian scene from both context and target views under the aligned poses for photometric supervision $\mathcal{L}_{photo}$. This objective explicitly encourages consistency between the reconstructed scene and the predicted camera poses, while ground-truth pose supervision $\mathcal{L}_{pose}$ anchors pose estimation and prevents mutually compensating errors between scene geometry and camera poses. \par

\textbf{Rendering-Consistent Instance Learning.}
For instance learning, we aim to group Gaussian primitives into instances based on their attached instance embeddings. A straightforward solution is to apply pixel-level contrastive learning to the pixel-aligned embeddings from the context views. However, enumerating pixel pairs is computationally expensive and may overemphasize redundant local variations. We therefore formulate direct supervision using instance centroids. From $N$ sampled context embeddings with corresponding instance labels $\{\mathbf e_n,y_n\}_{n=1}^{N}$, we compute an $\ell_2$-normalized centroid $\bar{\mathbf e}_k$ for each of the $K$ instances, where $\mathbf e_n$ and $y_n$ denote the normalized embedding and instance label, respectively. The direct objective pulls embeddings toward their corresponding centroids and pushes different centroids apart by a margin $m$:
\begin{equation}
\begin{aligned}
\mathcal L_{\mathrm{direct}}^{\mathrm{pull}}
&=
\frac{1}{N}\sum_{n=1}^{N}
\left\lVert
\mathbf e_n-\bar{\mathbf e}_{y_n}
\right\rVert_2, \\
\mathcal L_{\mathrm{direct}}^{\mathrm{push}}
&=
\frac{1}{K(K-1)}
\sum_{k\neq l}
\left[
m-
\left\lVert
\bar{\mathbf e}_k-\bar{\mathbf e}_l
\right\rVert_2
\right]_+ .
\end{aligned}
\end{equation}
We combine pull and push terms to form the direct objective:
\begin{equation}
\mathcal L_{\mathrm{direct}}
=
\lambda_{\mathrm{pull}}
\mathcal L_{\mathrm{direct}}^{\mathrm{pull}}
+
\lambda_{\mathrm{push}}
\mathcal L_{\mathrm{direct}}^{\mathrm{push}}.
\end{equation}

However, this context-space objective alone does not explicitly account for changes introduced by Gaussian rasterization, including visibility, occlusion, and alpha compositing. We therefore compute the assignment probability of Gaussian $g$ to instance $k$ as
\begin{equation}
p_{gk}
=
\operatorname{softmax}_{k}
\left(
-\frac{
\left\lVert
\mathbf e_g-\bar{\mathbf e}_k
\right\rVert_2
}{\tau}
\right),
\end{equation}
where $\tau$ is a temperature parameter. The probabilities over all $K$ instances form the assignment vector $\mathbf p_g=(p_{g1},\ldots,p_{gK})$. During training, $\mathbf p_g$ is temporarily attached to Gaussian $g$ as an auxiliary renderable attribute and rasterized into both context and target views. Each assignment channel represents a distinct object instance rather than a semantic category. We supervise the rendered partitions against ground-truth instance masks using a dice loss $\mathcal L_{\mathrm{render}}^{\mathrm{dice}}$~\cite{dice} for region-level alignment and a focal loss $\mathcal L_{\mathrm{render}}^{\mathrm{focal}}$~\cite{focal} for ambiguous and low-confidence instance assignments. In parallel, we render the Gaussian embeddings into the same views and sample $N_{\mathrm{r}}$ rendered embeddings. Each sampled embedding is pulled toward its corresponding context centroid computed before rendering, while different rendered centroids are pushed by a margin $m$:
\begin{equation}
\begin{aligned}
\mathcal L_{\mathrm{render}}^{\mathrm{pull}}
&=
\frac{1}{N_{\mathrm{r}}}
\sum_{r=1}^{N_{\mathrm{r}}}
\left\lVert
\widehat{\mathbf e}_r
- 
\bar{\mathbf e}_{y_r}
\right\rVert_2, \\
\mathcal L_{\mathrm{render}}^{\mathrm{push}}
&=
\frac{1}{K(K-1)}
\sum_{k\neq l}
\left[
m-
\left\lVert
\widehat{\bar{\mathbf e}}_k
-
\widehat{\bar{\mathbf e}}_l
\right\rVert_2
\right]_+ .
\end{aligned}
\end{equation}
where $\widehat{\mathbf e}_r$ and $y_r$ denote the normalized rendered embedding and its ground-truth instance label, respectively. We combine the dice and focal losses on the rendered partitions with the pull and push losses on the rendered embeddings to form the rendering objective:
\begin{equation}
\begin{aligned}
\mathcal L_{\mathrm{render}}
&=
\lambda_{\mathrm{dice}}
\mathcal L_{\mathrm{render}}^{\mathrm{dice}}
+
\lambda_{\mathrm{focal}}
\mathcal L_{\mathrm{render}}^{\mathrm{focal}}
\\
&\quad+
\lambda_{\mathrm{pull}}
\mathcal L_{\mathrm{render}}^{\mathrm{pull}}
+
\lambda_{\mathrm{push}}
\mathcal L_{\mathrm{render}}^{\mathrm{push}}.
\end{aligned}
\end{equation}
The complete instance-learning objective is
\begin{equation}
\mathcal L_{\mathrm{ins}}
=
\lambda_{\mathrm{direct}}
\mathcal L_{\mathrm{direct}}
+
\lambda_{\mathrm{render}} \mathcal L_{\mathrm{render}}.
\end{equation}
The direct objective supervises instance discrimination in the context-view embedding space, while the rendering objective further enforces instance consistency in 3D space. Collectively, these objectives facilitate instance embedding learning and enforce consistency across context and target views. \par

\textbf{Semantic Aggregation and Propagation.}
After obtaining discriminative Gaussian instance embeddings, we jointly cluster the embeddings from all context views using HDBSCAN~\cite{hdbscan} at inference time. Since each embedding is attached to a pixel-aligned Gaussian, the resulting clusters directly define cross-view consistent 3D Gaussian groups. We then extract dense semantic features $\mathbf F_v$ from each context image using a frozen 2D open-vocabulary model~\cite{lseg}. For instance group $k$, its semantic prototype is obtained by pooling the features within its context-view partitions:
\begin{equation}
\mathbf f_k^{\mathrm{sem}}
=
\frac{
\displaystyle\sum_{v\in\mathcal C}\sum_u
M_{v,k}(u)\mathbf F_v(u)
}{
\displaystyle\sum_{v\in\mathcal C}\sum_u
M_{v,k}(u)
},
\end{equation}
where $M_{v,k}(u)\in\{0,1\}$ denotes the instance mask obtained by mapping Gaussian group $k$ to its corresponding pixels in context view $v$ through the pixel-Gaussian correspondence, and $\mathbf f_k^{\mathrm{sem}}$ is the corresponding semantic prototype. The instance assignments are subsequently attached to the pixel-aligned Gaussians and rendered into target views to obtain target-view instance partitions. Semantic features are propagated by combining the rendered partition weights with their corresponding prototypes. Consequently, high-dimensional semantic features are stored once per instance rather than for every Gaussian. The low-dimensional embeddings organize Gaussians into instance groups, while the high-dimensional semantics are aggregated and propagated at the instance level. \par

\textbf{Staged Curriculum Training.}
We adopt a three-stage curriculum training strategy. In Stage 1, we freeze the instance head and optimize the shared encoder, geometry decoder, and Gaussian decoder for scene reconstruction. The reconstruction objective $\mathcal L_{\mathrm{rec}}$ combines $\ell_1$ and SSIM photometric losses with an additional LPIPS loss, as well as pose, depth, point-map, and normal supervision. In Stage 2, we freeze the shared encoder, Geometry Decoder, and attribute head. We only train the instance head using $\mathcal L_{\mathrm{ins}}$. In Stage 3, we jointly fine-tune the full model with
\begin{equation}
\mathcal L_{\mathrm{total}}
=
\mathcal L_{\mathrm{rec}}
+
\mathcal L_{\mathrm{ins}},
\end{equation}
This curriculum training establishes stable geometry and appearance before introducing instance learning, thereby improving training stability and preventing early optimization collapse. Joint optimization in the final stage further facilitates coordinated adaptation of the Geometry and Gaussian decoders. In addition, the staged training also supports instance learning without ground-truth instance labels. When such labels are unavailable, we first extract instance masks independently from each view using SAM2~\cite{sam2}. The depth and camera geometry learned in Stage 1 are then used to project these masks across views, discard geometrically inconsistent correspondences, and associate the remaining masks according to their projected overlap. This process establishes consistent identities across the per-view masks and provides pseudo-labels for instance supervision in Stage 2.\par

\subsection{Referential Scene Reasoning Framework}
\label{sec:rsrf}

The Referential Scene Reasoning Framework (RSRF) enables structured referential reasoning over instance-grouped 3D scenes and achieves 3D referring segmentation. Its design includes 3D scene graph construction, expression-guided instance retrieval, and VLM-based reasoning with structured relational evidence and multi-view observations. \par

\textbf{3D Scene Graph over Instance Groups.}
We organize the reconstructed scene into an instance-level relational graph $\mathcal G=(\mathcal N,\mathcal E)$, where each node corresponds to a cross-view consistent 3D Gaussian group and each ordered edge encodes the spatial relation of one group to another:
\begin{equation}
\begin{aligned}
\mathbf n_k
&=
\left(
\mathbf f_k^{\mathrm{sem}},
\mathbf g_k^{3\mathrm D},
\left\{
\mathbf g_{v,k}^{2\mathrm D},
\nu_{v,k}
\right\}_{v\in\mathcal C}
\right),\\
\mathbf e_{kl}
&=
\left(
d_{kl}^{3\mathrm D},
\Delta h_{kl},
r_{kl},
\left\{
\mathbf r_{v,kl}^{2\mathrm D}
\right\}_{v\in\mathcal C}
\right).
\end{aligned}
\end{equation}
Here, $\mathbf f_k^{\mathrm{sem}}$ denotes the semantic prototype of node $k$, while $\mathbf g_k^{3\mathrm D}$ summarizes its 3D center, spatial extent, and volume. The view-dependent attributes $\mathbf g_{v,k}^{2\mathrm D}$ and $\nu_{v,k}$ describe its projected geometry and visibility in view $v$, respectively. For an ordered node pair $(k,l)$, the edge attributes encode their 3D distance $d_{kl}^{3\mathrm D}$, relative height $\Delta h_{kl}$, size ratio $r_{kl}$, and view-dependent 2D relations $\mathbf r_{v,kl}^{2\mathrm D}$, including projected distance, direction, and overlap. Each node additionally retains references to its associated masks and visual observations, linking the structured graph representation to the corresponding multi-view evidence. \par

\textbf{Instance Candidate Retrieval.}
Given a referring expression, RSRF decomposes it into target concepts, target cardinality, anchor instances, spatial relations, and visual attributes. These query components are then used to retrieve relevant nodes from the scene graph. Semantic relevance is measured by comparing the target and anchor concepts with the instance-level semantic prototypes and open-vocabulary labels, while geometric relevance evaluates whether the spatial relations between candidate nodes and inferred anchors are compatible with the expression. Cross-view visibility is also considered to favor candidates supported by sufficient visual evidence. The resulting semantic, geometric, and visibility scores are combined to rank the graph nodes, and the highest-scoring nodes are retained for subsequent reasoning. This stage reduces the VLM reasoning space while preserving candidates that satisfy the principal semantic and spatial constraints of the expression. \par

\textbf{VLM Reasoning with Structured Evidence.}
For each retrieved candidate, we construct a node-centric evidence bundle containing its semantic and geometric attributes, relations to neighboring nodes, cross-view visibility, and multi-view observations. The visual evidence includes global scene overviews, highlighted instance masks, object-centric crops, and local contexts that expose both appearance and surrounding instances. A vision-language model (VLM) reasons over the referring expression, structured graph evidence, and multi-view observations to identify the target 3D instance entities from the retrieved candidates, consistent with the inferred target cardinality. By restricting the VLM to selecting among the retrieved candidate nodes, RSRF formulates referential grounding as a closed-set selection problem over reconstructed 3D instances. Since each selected node maintains explicit correspondences to its observations and masks across views, the grounded 3D entities can be directly mapped to cross-view consistent segmentations without additional mask prediction. \par

\begin{table*}[t!]
\centering
\scriptsize
\caption{
Quantitative comparison on ScanNet with two input views, following the benchmark established by LSM~\cite{lsm} and Uni3R~\cite{uni3r}.
}
\label{tab:1_main}
\resizebox{\textwidth}{!}{
\begin{tabular}{l|c|cccc|ccccc}
\toprule
\multirow{2}{*}{Method}
& \multirow{2}{*}{Venue \& Year}
& \multicolumn{4}{c|}{Source View}
& \multicolumn{5}{c}{Target View} \\
\cmidrule(lr){3-6}
\cmidrule(lr){7-11}
& 
& mIoU$\uparrow$ & Acc.$\uparrow$ & rel$\downarrow$ & $\tau\uparrow$
& mIoU$\uparrow$ & Acc.$\uparrow$ & PSNR$\uparrow$ & SSIM$\uparrow$ & LPIPS$\downarrow$ \\
\midrule
LSeg~\cite{lseg}         & ICLR 2022 & 0.4701 & 0.7891 & --    & --    & 0.4819 & 0.7927 & --     & --     & --     \\
NeRF-DEF~\cite{dff}     & NeurIPS 2022 & 0.4540 & 0.7173 & 27.68 & 9.61  & 0.4037 & 0.6755 & 19.86  & 0.6650 & 0.3629 \\
Feature 3DGS~\cite{feature3dgs} & CVPR 2024 & 0.4453 & 0.7276 & 12.95 & 21.07 & 0.4223 & 0.7174 & 24.49  & 0.8132 & 0.2293 \\
\midrule
PixelSplat~\cite{pixelsplat}   & CVPR 2024 & --     & --     & --    & --    & --     & --     & 24.89  & 0.8392 & 0.1641 \\
LSM~\cite{lsm}          & NeurIPS 2024 & 0.5034 & 0.7740 & 3.38 & \underline{67.77} & 0.5078 & 0.7686 & 24.39 & 0.8072 & 0.2506 \\
AnySplat~\cite{anysplat}     & SIGGRAPH Asia 2025 & --     & --     & 6.35  & 47.57 & --     & --     & 22.08  & 0.8118 & 0.2480 \\
IGGT~\cite{iggt}         & ICLR 2026 & \underline{0.5626} & 0.8145 & \underline{3.15}  & 66.94 & --     & --     & --     & --     & --     \\
Uni3R~\cite{uni3r}        & CVPR 2026 & 0.5403 & \underline{0.8255} & 3.87 & 61.37 & \underline{0.5584} & \underline{0.8268} & \underline{25.53} & \underline{0.8727} & \underline{0.1380} \\
Ours         & --        & \textbf{0.5905} & \textbf{0.8294} & \textbf{2.14} & \textbf{82.23} & \textbf{0.5989} & \textbf{0.8302} & \textbf{26.23} & \textbf{0.8948} & \textbf{0.1230} \\
\bottomrule
\end{tabular}
}
\end{table*}

\begin{table*}[t!]
\centering
\scriptsize
\caption{
Quantitative comparison on ScanNet with varying numbers of input views for novel-view semantic segmentation and synthesis.
}
\label{tab:2_mv}
\resizebox{\textwidth}{!}{
\begin{tabular}{l|ccccc|ccccc}
\toprule

\multirow{2}{*}{Method}
& \multicolumn{5}{c|}{4 Views}
& \multicolumn{5}{c}{6 Views} \\
\cmidrule(lr){2-6}
\cmidrule(lr){7-11}
& mIoU$\uparrow$
& Acc.$\uparrow$
& PSNR$\uparrow$
& SSIM$\uparrow$
& LPIPS$\downarrow$
& mIoU$\uparrow$
& Acc.$\uparrow$
& PSNR$\uparrow$
& SSIM$\uparrow$
& LPIPS$\downarrow$ \\
\midrule

LSeg~\cite{lseg}
& 0.5002 & 0.7982 & -- & -- & --
& 0.5381 & 0.8077 & -- & -- & -- \\

IGGT~\cite{iggt}
& 0.5247 & 0.7911 & -- & -- & --
& \underline{0.5799} & 0.8131 & -- & -- & -- \\

AnySplat~\cite{anysplat}
& -- & -- & 21.82 & 0.8204 & 0.1415
& -- & -- & 22.28 & 0.8093 & 0.1509 \\

Feature 3DGS~\cite{feature3dgs}
& 0.5192 & 0.7250 & 23.78 & 0.7779 & 0.2661
& 0.5461 & 0.7399 & 24.36 & 0.7905 & 0.2356 \\

LSM~\cite{lsm}
& 0.4949 & 0.8002 & 22.85 & 0.8066 & 0.1971
& 0.5323 & 0.8076 & 20.80 & 0.7986 & 0.2056 \\

Uni3R~\cite{uni3r}
& \underline{0.5443} & \underline{0.8118} & \underline{25.32} & \underline{0.8438} & \underline{0.1340}
& 0.5794 & \underline{0.8186} & \underline{25.34} & \underline{0.8441} & \underline{0.1422} \\

Ours
& \textbf{0.5868} & \textbf{0.8277} & \textbf{27.14} & \textbf{0.9015} & \textbf{0.1083}
& \textbf{0.6513} & \textbf{0.8467} & \textbf{26.51} & \textbf{0.8890} & \textbf{0.1198} \\

\midrule

& \multicolumn{5}{c|}{8 Views}
& \multicolumn{5}{c}{16 Views} \\
\cmidrule(lr){2-6}
\cmidrule(lr){7-11}
& mIoU$\uparrow$
& Acc.$\uparrow$
& PSNR$\uparrow$
& SSIM$\uparrow$
& LPIPS$\downarrow$
& mIoU$\uparrow$
& Acc.$\uparrow$
& PSNR$\uparrow$
& SSIM$\uparrow$
& LPIPS$\downarrow$ \\

LSeg~\cite{lseg}
& 0.5466 & 0.8175 & -- & -- & --
& 0.5387 & 0.8201 & -- & -- & -- \\

IGGT~\cite{iggt}
& 0.5719 & 0.8346 & -- & -- & --
& \underline{0.6002} & \underline{0.8596} & -- & -- & -- \\

AnySplat~\cite{anysplat}
& -- & -- & 21.98 & 0.7724 & 0.1722
& -- & -- & 18.92 & 0.7105 & 0.2372 \\

Feature 3DGS~\cite{feature3dgs}
& 0.5810 & 0.7745 & 24.19 & 0.7929 & 0.2239
& 0.5046 & 0.8289 & \underline{23.78} & \underline{0.8184} & \underline{0.1812} \\

LSM~\cite{lsm}
& 0.5567 & 0.8202 & 20.63 & 0.7888 & 0.2125
& 0.5621 & 0.8319 & 19.07 & 0.7813 & 0.2262 \\

Uni3R~\cite{uni3r}
& \underline{0.5880} & \underline{0.8359} & \underline{24.99} & \underline{0.8318} & \underline{0.1580}
& 0.5813 & 0.8484 & 23.35 & 0.7827 & 0.2089 \\

Ours
& \textbf{0.6623} & \textbf{0.8669} & \textbf{26.01} & \textbf{0.8740} & \textbf{0.1303}
& \textbf{0.6880} & \textbf{0.8893} & \textbf{25.71} & \textbf{0.8629} & \textbf{0.1495} \\

\bottomrule
\end{tabular}
}
\end{table*}

\section{Experiments}

\subsection{Experimental Setup}

\textbf{Implementation Details.}
We implement GroupForward and RSRF in PyTorch. For GroupForward, we use VGGT~\cite{vggt} as its multi-view encoder and initialize both the encoder and geometry decoder with pretrained VGGT weights. The remaining prediction heads are randomly initialized. We use gsplat~\cite{gsplat} for differentiable Gaussian rasterization. We set the instance embedding dimension to $d_e=8$, substantially smaller than the 512-dimensional LSeg features and the 64- or higher-dimensional embeddings used by existing baselines~\cite{feature3dgs,lsm,uni3r}. The learning rates are $10^{-6}$ for the encoder and $10^{-5}$ for the decoders. We empirically set $\lambda_{\mathrm{pull}}=\lambda_{\mathrm{push}}=\lambda_{\mathrm{dice}}=\lambda_{\mathrm{focal}}=0.5$, $m=1.0$, and $\lambda_{\mathrm{direct}}=\lambda_{\mathrm{render}}=1.0$. Following the staged curriculum in Sec.~\ref{sec:GroupForward}, we train each stage for approximately 50k iterations in the two-view setting, which takes approximately 30 hours on 8 $\times$ NVIDIA A100 GPUs (80GB). Since GroupForward naturally supports a variable number of input views, we further fine-tune the two-view model for 50k iterations using 2--24 input views for multi-view settings. Following LSM and Uni3R~\cite{lsm,uni3r}, we also adopt LSeg~\cite{lseg} as the semantic model. GroupForward aggregates and propagates the extracted LSeg features at the instance level. For RSRF, we adopt Qwen3-VL-8B-Instruct~\cite{qwen3vl} as the VLM and deploy it with vLLM~\cite{vllm}. To bound the context length, we rank scene-graph nodes by semantic similarity, geometric relevance, and multi-view visibility, and retain the top-$K$ nodes ($K=8$) as candidates for VLM reasoning. We run RSRF on a single NVIDIA A100 GPU (80GB). \par

\textbf{Datasets and Evaluation Protocol.}
We follow the two-input-view ScanNet benchmark and evaluation protocol established by LSM~\cite{lsm} and Uni3R~\cite{uni3r}. We train GroupForward on the ScanNet~\cite{scannet} training set, with strictly non-overlapping training and test scenes. We further evaluate more input views on ScanNet and zero-shot generalization on Replica~\cite{replica}. We report mean intersection over union (mIoU) and pixel accuracy (Acc.) for semantic reconstruction via input- and novel-view semantic segmentation, absolute relative error (rel) and threshold accuracy ($\tau$) for geometry estimation, and peak signal-to-noise ratio (PSNR), structural similarity index measure (SSIM)~\cite{ssim}, and learned perceptual image patch similarity (LPIPS)~\cite{lpips} for novel-view synthesis. For referential scene reasoning, we construct referring expressions paired with target instances and evaluate novel-view referring segmentation against the corresponding ground-truth masks. We report mIoU averaged over all queries and novel views. \par

\textbf{Baselines.}
We primarily compare our method with semantic 3D reconstruction methods, including feed-forward 3DGS methods~\cite{lsm, uni3r}, per-scene optimized methods~\cite{dff, feature3dgs}, and pointmap-based semantic reconstruction method~\cite{iggt}. We further include semantic-only~\cite{lseg} and reconstruction-only methods~\cite{pixelsplat, anysplat} for comprehensive evaluation. We use the published results for the existing benchmark. For additional comparisons, we evaluate the baselines using their official implementations and open-source pretrained weights. \par

\begin{table}[t!]
\centering
\footnotesize
\caption{
Zero-shot generalization from ScanNet to Replica. \\
Results are averaged over 2, 4, 6, and 8 input views.
}
\label{tab:3_replica_zeroshot}
\begin{tabular*}{\columnwidth}{@{\extracolsep{\fill}}lccccc}
\toprule
~Method & mIoU$\uparrow$ & Acc.$\uparrow$ & PSNR$\uparrow$ & SSIM$\uparrow$ & LPIPS$\downarrow$ \\
\midrule
~LSeg~\cite{lseg} & 0.4690 & \underline{0.8410} & -- & -- & -- \\
~LSM~\cite{lsm} & 0.4341 & 0.8233 & 16.73 & 0.6539 & 0.2814 \\
~Uni3R~\cite{uni3r} & \underline{0.4754} & 0.8351 & \underline{23.03} & \underline{0.7880} & \underline{0.1276} \\
~Ours & \textbf{0.5633} & \textbf{0.8769} & \textbf{27.26} & \textbf{0.8790} & \textbf{0.1226} \\
\bottomrule
\end{tabular*}
\end{table}

\begin{table}[t!]
\centering
\footnotesize
\caption{
Comparison of referential scene reasoning capabilities with novel-view referring segmentation performance.
}
\label{tab:4_reasoning_table}
\begin{tabular*}{\columnwidth}{@{\extracolsep{\fill}}lccc}
\toprule
~Method
& Language Understanding
& 3D Output
& mIoU$\uparrow$ \\
\midrule
~Uni3R~\cite{uni3r}
& Phrase / Category
& $\checkmark$
& 0.2538 \\

~Sa2VA~\cite{sa2va}
& Referential
& $\times$
& 0.5012 \\

~\textbf{Ours}
& Referential
& $\checkmark$
& \textbf{0.6496} \\
\bottomrule
\end{tabular*}
\end{table}

\begin{table}[t!]
\centering
\footnotesize
\caption{
Comparison of different semantic models for GroupForward.
}
\label{tab:5_semantic_model_ablation}
\begin{tabular*}{0.7\columnwidth}{@{\extracolsep{\fill}}lcc}
\toprule
~Semantic Model & mIoU$\uparrow$ & Acc.$\uparrow$ \\
\midrule
~LSeg~\cite{lseg} (default) & 0.5905 & 0.8294 \\
~OpenSeg~\cite{openseg}        & 0.4536 & 0.6292 \\
~CLIP~\cite{clip}           & 0.1630 & 0.3759 \\
\bottomrule
\end{tabular*}
\end{table}

\begin{figure*}[t]
    \centering
    \includegraphics[width=0.98\textwidth]{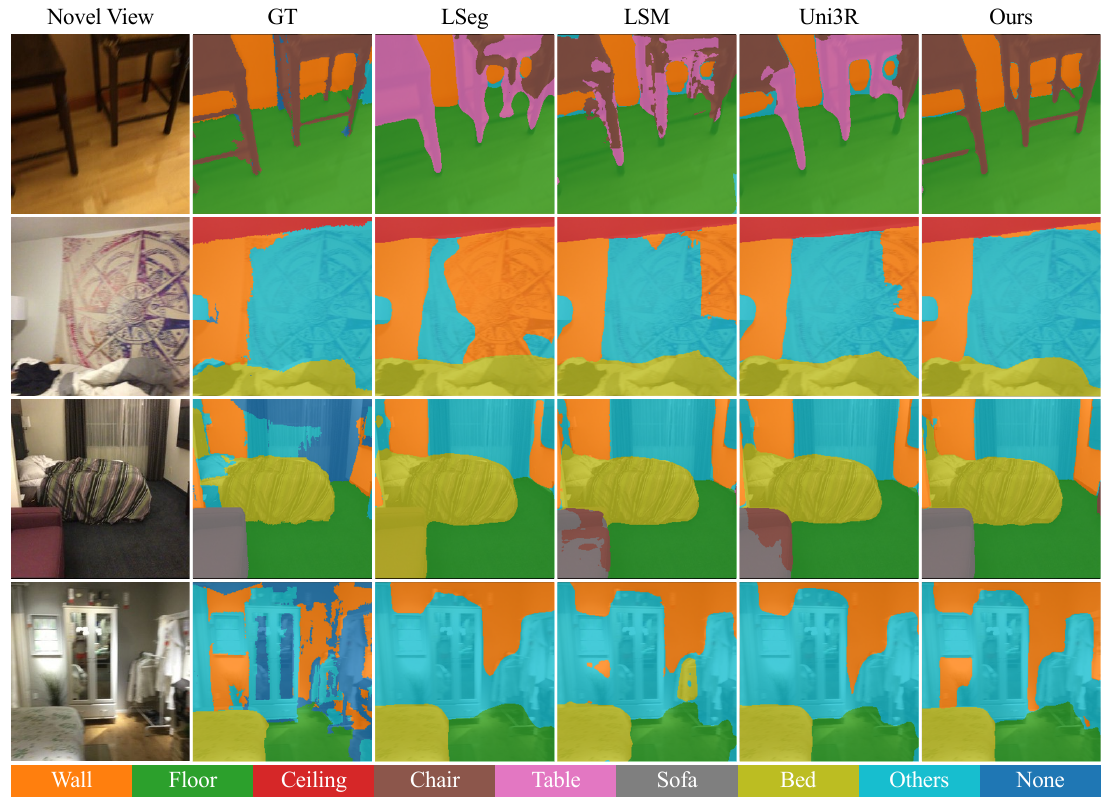}
    \caption{Qualitative comparisons of novel-view open-vocabulary segmentation on ScanNet. Our GroupForward produces more complete semantic regions with cleaner object boundaries and fewer cross-category inconsistencies than existing methods.}
    \label{fig:sem}
\end{figure*}

\subsection{Comparison on Semantic 3D Reconstruction}

\textbf{Quantitative Results on ScanNet.} Table~\ref{tab:1_main} compares the reconstruction performance of different methods on the existing two-input-view ScanNet benchmark~\cite{lsm,uni3r}. GroupForward achieves the best performance across all reported semantic, geometric, and appearance metrics. Unlike LSM and Uni3R, which attach high-dimensional semantic features to each Gaussian, GroupForward uses compact instance embeddings and propagates semantics through instance-level prototypes. Although IGGT~\cite{iggt} also learns instance-level features, it does not support novel-view synthesis and underperforms GroupForward on source-view semantic metrics. The consistent improvements demonstrate the effectiveness of our instance-grouped formulation in improving semantic reconstruction while preserving strong geometry and appearance quality. \par

\textbf{Results with More Input Views.} Table~\ref{tab:2_mv} reports results using 4, 6, 8, and 16 input views on ScanNet with our multi-view model. GroupForward consistently achieves the best performance across all input-view settings in both novel-view semantic segmentation and novel-view synthesis. These results demonstrate that the advantages of our instance-grouped formulation extend to multi-view settings. \par 

\textbf{Zero-Shot Generalization.} Table~\ref{tab:3_replica_zeroshot} evaluates zero-shot generalization across datasets, where our multi-view model trained on ScanNet is evaluated on Replica without fine-tuning. Results are averaged over settings with 2, 4, 6, and 8 input views. Our GroupForward achieves the best performance across all reported semantic and appearance metrics, consistently outperforming existing methods. These results demonstrate the strong cross-dataset generalization of our method to unseen scenes. \par

\begin{figure*}[t]
    \centering
    \includegraphics[width=0.98\textwidth]{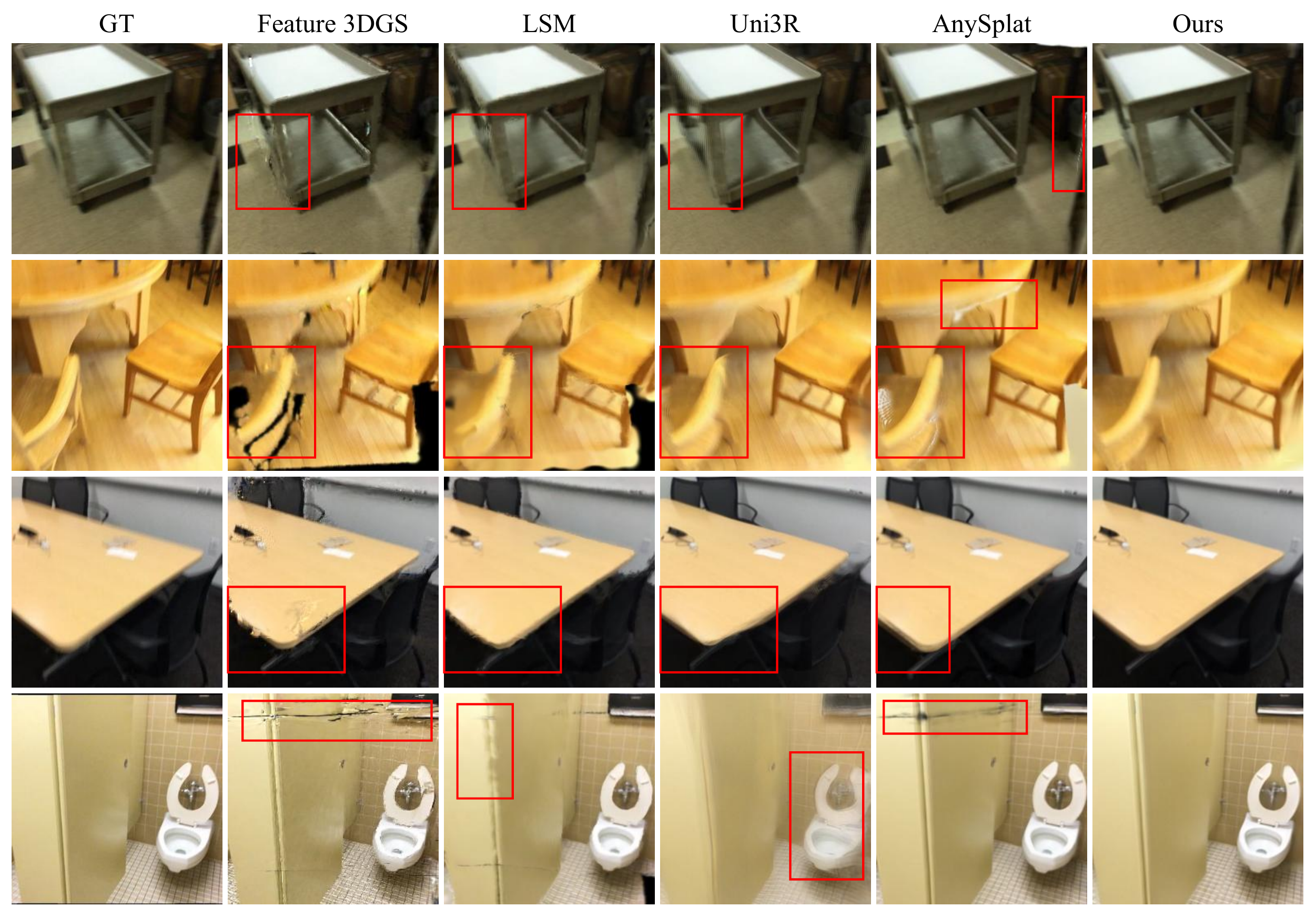}
    \caption{Qualitative comparison of novel-view synthesis on ScanNet. Red boxes highlight representative rendering artifacts in competing methods, while our GroupForward better preserves scene structure and fine-grained visual details.}
    \label{fig:nvs}
\end{figure*}

\textbf{Qualitative Results.} Fig.~\ref{fig:sem} and Fig.~\ref{fig:nvs} present qualitative comparisons of novel-view semantic segmentation and novel-view synthesis, respectively. GroupForward produces more complete semantic regions with clearer object boundaries and fewer cross-category inconsistencies in novel views. It also better preserves scene structure and visual details in the synthesized novel views. These observations are consistent with the quantitative improvements and show that instance-level aggregation promotes view-consistent semantic propagation without compromising rendering quality. \par

\subsection{Referential Scene Reasoning}

\begin{figure*}[t]
    \centering
    \includegraphics[width=0.98\textwidth]{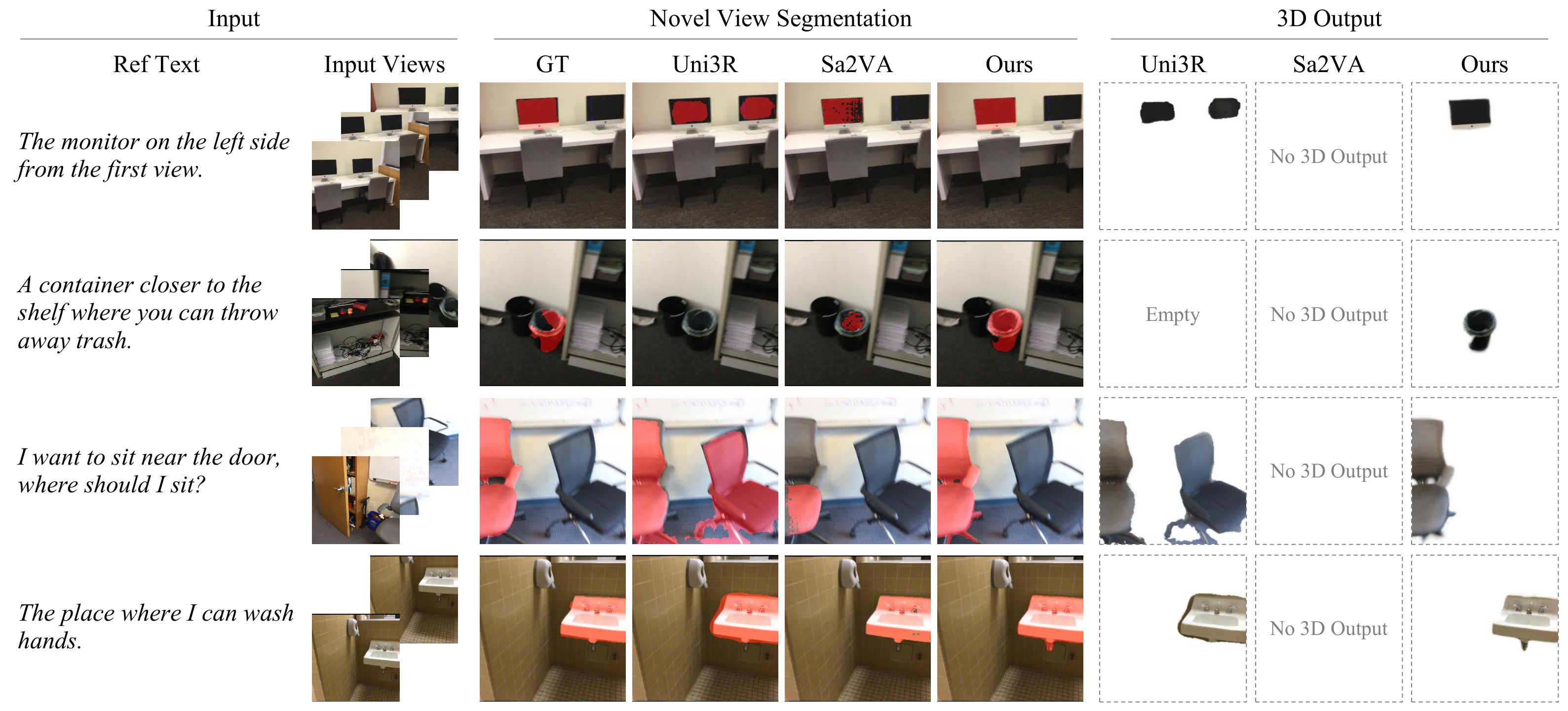}
    \caption{Qualitative comparisons of referential scene reasoning. Our method correctly grounds target instances with accurate novel-view segmentation and corresponding 3D outputs, while competing methods fail to ground the target instances or lack native 3D outputs.}
    \label{fig:reasoning}
\end{figure*}

Beyond semantic 3D reconstruction, we further evaluate the proposed Referential Scene Reasoning Framework (RSRF) built upon GroupForward through both quantitative and qualitative comparisons. As summarized in Table~\ref{tab:4_reasoning_table}, existing methods face inherent limitations in this task. Semantic reconstruction methods such as Uni3R~\cite{uni3r} rely on semantic similarity matching and therefore support only phrase- or category-level language queries, failing to handle complex referring expressions involving spatial relations and contextual constraints. Conversely, referring segmentation models that couple a segmentation model with a VLM, such as Sa2VA~\cite{sa2va}, can interpret complex referential language but do not support novel-view synthesis. We therefore directly take the novel views as input to Sa2VA for evaluation. Moreover, Sa2VA does not natively produce 3D outputs, resulting in inconsistent segmentation results across views. In contrast, RSRF grounds complex referring expressions to the cross-view consistent 3D instance groups reconstructed by GroupForward, jointly supporting complex referential understanding, view-consistent segmentation, and native 3D output. Consistent with these capabilities, RSRF substantially outperforms both baselines on novel-view referring segmentation. \par

Fig.~\ref{fig:reasoning} presents qualitative comparisons. In the first three rows, which involve complex referring expressions, our method correctly identifies the target instances, whereas the compared methods either identify incorrect objects or fail to produce valid segmentations. In the relatively simple case shown in the fourth row, our method still yields higher-quality novel-view segmentation together with a consistent 3D output, while Sa2VA, despite achieving comparable 2D accuracy, cannot provide a 3D result. These results demonstrate the advantage of our instance-grounded reasoning framework for referential scene reasoning over multi-view inputs. \par

\subsection{Compatibility with Different Semantic Models} As shown in Table~\ref{tab:5_semantic_model_ablation}, our GroupForward is compatible with different semantic models, including LSeg~\cite{lseg}, OpenSeg~\cite{openseg}, and CLIP~\cite{clip}. Following the established benchmark~\cite{lsm, uni3r}, semantic quality is evaluated using 2D multiclass segmentation including an explicit \textit{other} class. LSeg achieves the best performance among the evaluated models. OpenSeg is less effective at modeling the catch-all \textit{other} class, resulting in lower overall scores, while CLIP is less suitable for dense pixel-level alignment and fine-grained class discrimination due to its global image-text training objective. Since GroupForward aggregates dense semantic features into instance prototypes, feature quality directly affects semantic performance. We therefore use LSeg as the default semantic model, which is also consistent with  LSM~\cite{lsm} and Uni3R~\cite{uni3r} for a fair comparison. \par

\begin{table*}[t!]
\centering
\caption{Ablation studies of GroupForward on ScanNet. Starting from the default full model, we remove or replace key components to assess the contributions of individual design choices and the effectiveness of pseudo-label supervision.}
\label{tab:6_ablation_table}
\scriptsize
\resizebox{0.7\textwidth}{!}{
\begin{tabular}{clccccc}
\toprule
\# & Method 
& mIoU$\uparrow$ 
& Acc.$\uparrow$ 
& PSNR$\uparrow$ 
& SSIM$\uparrow$ 
& LPIPS$\downarrow$ \\
\midrule

1 & Ours (default)
& 0.5989 & 0.8302 & 26.23 & 0.8948 & 0.1230 \\

\midrule
\addlinespace[3pt]
\multicolumn{7}{c}{\textit{Semantic ablation}} \\
\addlinespace[3pt]
2 & w/o $\mathcal L_{\mathrm{render}}$
& 0.5772 & 0.8131 & 26.17 & 0.8924 & 0.1237 \\
3 & w/o instance embedding
& 0.5479 & 0.8096 & 26.19 & 0.8929 & 0.1234 \\

\midrule
\addlinespace[3pt]
\multicolumn{7}{c}{\textit{Rendering ablation}} \\
\addlinespace[3pt]
4 & w/o self-rendering
& 0.5950 & 0.8297 & 26.11 & 0.8917 & 0.1267 \\

\midrule
\addlinespace[3pt]
\multicolumn{7}{c}{\textit{Training ablation}} \\
\addlinespace[3pt]
5 & w/o curriculum training
& 0.4669 & 0.7486 & 18.92 & 0.6733 & 0.3122 \\

\midrule
\addlinespace[3pt]
\multicolumn{7}{c}{\textit{Pseudo-label ablation}} \\
\addlinespace[3pt]
6 & w/ VGGT
& 0.5879 & 0.8278 & 25.97 & 0.8862 & 0.1287 \\
7 & w/ SAM2
& 0.5649 & 0.8221 & 26.16 & 0.8911 & 0.1287 \\

\bottomrule
\end{tabular}
}
\end{table*}

\subsection{Ablation Studies}
We conduct ablation studies on ScanNet to evaluate the contributions of key components and the effectiveness of pseudo-label supervision. The results are reported in Table~\ref{tab:6_ablation_table}.

\textbf{Instance Components.}
For instance- and semantic-related components, we separately ablate the loss design and the instance-grouped representation. Removing the rendering-based objective $\mathcal L_{\mathrm{render}}$ and retaining only the context-view objective $\mathcal L_{\mathrm{direct}}$ (w/o $\mathcal L_{\mathrm{render}}$, row 2) leads to a clear drop in mIoU, validating the importance of rendering-level supervision for instance learning. We further remove the instance embedding representation entirely (w/o instance embedding, row 3) and follow LSM and Uni3R by attaching high-dimensional semantic features to the Gaussians and supervising the rendered feature maps. The larger performance degradation highlights the advantage of our instance-grouped representation over direct high-dimensional semantic feature rendering. Meanwhile, both variants exhibit only marginal changes in appearance quality, indicating that these components mainly contribute to semantic reconstruction. \par

\textbf{Gauge-Aligned Self-Rendering.}
Removing gauge-aligned self-rendering (w/o self-rendering, row 4) degrades all appearance metrics, with lower PSNR and SSIM and higher LPIPS, while semantic performance decreases slightly. These results show that gauge-aligned self-rendering improves consistency between the reconstructed scene and predicted camera poses, thereby benefiting appearance reconstruction. \par

\textbf{Staged Curriculum Training.}
We remove the staged curriculum and optimize all components jointly from the beginning (w/o curriculum training, row 5). This variant causes a substantial degradation across both semantic and appearance metrics. Since instance supervision relies on sufficiently reliable geometry and poses, introducing it too early leads to unstable optimization. The result indicates that the staged curriculum is important for establishing a stable reconstruction before instance learning. \par

\textbf{Pseudo-Label Supervision.}
To evaluate the applicability of our method when ground-truth supervision is unavailable, we separately replace ground-truth geometric and instance supervision with predicted pseudo-labels and retrain the corresponding models. Using VGGT predictions for geometric supervision (w/ VGGT, row 6) results in only a moderate degradation, suggesting that predicted geometry remains sufficiently reliable for reconstruction and instance learning. Replacing ground-truth instance labels with SAM2 pseudo masks (w/ SAM2, row 7) also preserves competitive semantic and appearance performance. Although pseudo-supervision introduces some degradation relative to the default setting, both variants still maintain strong performance against other methods in Table~\ref{tab:1_main}. This demonstrates that our method can effectively learn from both geometric and instance pseudo-labels, reducing the reliance on ground-truth annotations. \par

\section{Conclusion}

We introduced GroupForward, an instance-grouped feed-forward Gaussian splatting model for semantic 3D reconstruction from sparse, unposed, and uncalibrated multi-view images. Rather than attaching high-dimensional semantic features to each Gaussian, GroupForward learns compact embeddings that organize the primitives into cross-view consistent instance groups, while open-vocabulary semantics are aggregated and propagated at the instance level. Built on these instances, we further introduce the Referential Scene Reasoning Framework (RSRF), which constructs a 3D scene graph and employs a VLM with structured graph evidence and multi-view observations to identify target 3D entities from complex referring expressions. Experiments show that GroupForward consistently outperforms existing methods in semantic 3D reconstruction across different input-view settings and generalizes well to unseen scenes, while RSRF substantially improves referring segmentation with view-consistent and native 3D outputs. Ablation studies further validate the effectiveness of the proposed components. Overall, our instance-grouped formulation enables explicit instance discrimination in feed-forward semantic 3DGS and extends the language interaction capability from category- or phrase-level semantic querying to complex referential scene reasoning. \par


\bibliographystyle{IEEEtran}
\bibliography{IEEEabrv,main}

\begin{thebibliography}{10}
\providecommand{\url}[1]{#1}
\csname url@samestyle\endcsname
\providecommand{\newblock}{\relax}
\providecommand{\bibinfo}[2]{#2}
\providecommand{\BIBentrySTDinterwordspacing}{\spaceskip=0pt\relax}
\providecommand{\BIBentryALTinterwordstretchfactor}{4}
\providecommand{\BIBentryALTinterwordspacing}{\spaceskip=\fontdimen2\font plus
\BIBentryALTinterwordstretchfactor\fontdimen3\font minus \fontdimen4\font\relax}
\providecommand{\BIBforeignlanguage}[2]{{%
\expandafter\ifx\csname l@#1\endcsname\relax
\typeout{** WARNING: IEEEtran.bst: No hyphenation pattern has been}%
\typeout{** loaded for the language `#1'. Using the pattern for}%
\typeout{** the default language instead.}%
\else
\language=\csname l@#1\endcsname
\fi
#2}}
\providecommand{\BIBdecl}{\relax}
\BIBdecl

\bibitem{3dgs}
B.~Kerbl, G.~Kopanas, T.~Leimk{\"{u}}hler, and G.~Drettakis, ``3d gaussian splatting for real-time radiance field rendering,'' \emph{{ACM} Trans. Graph.}, vol.~42, no.~4, pp. 139:1--139:14, 2023.

\bibitem{langsplat}
M.~Qin, W.~Li, J.~Zhou, H.~Wang, and H.~Pfister, ``Langsplat: 3d language gaussian splatting,'' in \emph{Proceedings of the IEEE/CVF Conference on Computer Vision and Pattern Recognition (CVPR)}, 2024, pp. 20\,051--20\,060.

\bibitem{feature3dgs}
S.~Zhou, H.~Chang, S.~Jiang, Z.~Fan, Z.~Zhu, D.~Xu, P.~Chari, S.~You, Z.~Wang, and A.~Kadambi, ``Feature 3dgs: Supercharging 3d gaussian splatting to enable distilled feature fields,'' in \emph{Proceedings of the IEEE/CVF Conference on Computer Vision and Pattern Recognition (CVPR)}, 2024, pp. 21\,676--21\,685.

\bibitem{legaussians}
J.-C. Shi, M.~Wang, H.-B. Duan, and S.-H. Guan, ``Language embedded 3d gaussians for open-vocabulary scene understanding,'' in \emph{Proceedings of the IEEE/CVF Conference on Computer Vision and Pattern Recognition (CVPR)}, 2024, pp. 5333--5343.

\bibitem{opengaussian}
Y.~Wu, J.~Meng, H.~Li, C.~Wu, Y.~Shi, X.~Cheng, C.~Zhao, H.~Feng, E.~Ding, J.~Wang \emph{et~al.}, ``Opengaussian: Towards point-level 3d gaussian-based open vocabulary understanding,'' \emph{Advances in Neural Information Processing Systems (NeurIPS)}, vol.~37, pp. 19\,114--19\,138, 2024.

\bibitem{langsplatv2}
W.~Li, Y.~Zhao, M.~Qin, Y.~Liu, Y.~Cai, C.~Gan, and H.~Pfister, ``Langsplatv2: High-dimensional 3d language gaussian splatting with 450+ {FPS},'' in \emph{Advances in Neural Information Processing Systems (NeurIPS)}, 2025.

\bibitem{lsm}
Z.~Fan, J.~Zhang, W.~Cong, P.~Wang, R.~Li, K.~Wen, S.~Zhou, A.~Kadambi, Z.~Wang, D.~Xu \emph{et~al.}, ``Large spatial model: End-to-end unposed images to semantic 3d,'' \emph{Advances in Neural Information Processing Systems (NeurIPS)}, vol.~37, pp. 40\,212--40\,229, 2024.

\bibitem{uni3r}
X.~Sun, H.~Jiang, L.~Liu, S.~Nam, G.~Kang, X.~Wang, W.~Sui, Z.~Su, W.~Liu, X.~Wang \emph{et~al.}, ``Uni3r: Unified 3d reconstruction and semantic understanding via generalizable gaussian splatting from unposed multi-view images,'' in \emph{Proceedings of the IEEE/CVF Conference on Computer Vision and Pattern Recognition (CVPR)}, 2026, pp. 33\,280--33\,290.

\bibitem{sfm}
N.~Snavely, S.~M. Seitz, and R.~Szeliski, ``Photo tourism: Exploring photo collections in 3d,'' in \emph{ACM SIGGRAPH 2006 Papers}, 2006, pp. 835--846.

\bibitem{nerf}
B.~Mildenhall, P.~P. Srinivasan, M.~Tancik, J.~T. Barron, R.~Ramamoorthi, and R.~Ng, ``Nerf: Representing scenes as neural radiance fields for view synthesis,'' \emph{Communications of the ACM}, vol.~65, no.~1, pp. 99--106, 2021.

\bibitem{conceptgraphs}
Q.~Gu, A.~Kuwajerwala, S.~Morin, K.~M. Jatavallabhula, B.~Sen, A.~Agarwal, C.~Rivera, W.~Paul, K.~Ellis, R.~Chellappa, C.~Gan, C.~M. de~Melo, J.~B. Tenenbaum, A.~Torralba, F.~Shkurti, and L.~Paull, ``Conceptgraphs: Open-vocabulary 3d scene graphs for perception and planning,'' in \emph{International Conference on Robotics and Automation (ICRA)}.\hskip 1em plus 0.5em minus 0.4em\relax {IEEE}, 2024, pp. 5021--5028.

\bibitem{3dmem}
Y.~Yang, H.~Yang, J.~Zhou, P.~Chen, H.~Zhang, Y.~Du, and C.~Gan, ``3d-mem: 3d scene memory for embodied exploration and reasoning,'' in \emph{Proceedings of the IEEE/CVF Conference on Computer Vision and Pattern Recognition (CVPR)}, June 2025, pp. 17\,294--17\,303.

\bibitem{mtu3d}
Z.~Zhu, X.~Wang, Y.~Li, Z.~Zhang, X.~Ma, Y.~Chen, B.~Jia, W.~Liang, Q.~Yu, Z.~Deng, S.~Huang, and Q.~Li, ``Move to understand a 3d scene: Bridging visual grounding and exploration for efficient and versatile embodied navigation,'' in \emph{Proceedings of the IEEE/CVF International Conference on Computer Vision (ICCV)}, October 2025, pp. 8120--8132.

\bibitem{pixelnerf}
A.~Yu, V.~Ye, M.~Tancik, and A.~Kanazawa, ``pixelnerf: Neural radiance fields from one or few images,'' in \emph{Proceedings of the IEEE/CVF Conference on Computer Vision and Pattern Recognition (CVPR)}, 2021, pp. 4578--4587.

\bibitem{dust3r}
S.~Wang, V.~Leroy, Y.~Cabon, B.~Chidlovskii, and J.~Revaud, ``Dust3r: Geometric 3d vision made easy,'' in \emph{Proceedings of the IEEE/CVF Conference on Computer Vision and Pattern Recognition (CVPR)}, 2024, pp. 20\,697--20\,709.

\bibitem{mast3r}
V.~Leroy, Y.~Cabon, and J.~Revaud, ``Grounding image matching in 3d with mast3r,'' in \emph{European Conference on Computer Vision (ECCV)}, 2024, pp. 71--91.

\bibitem{pixelsplat}
D.~Charatan, S.~L. Li, A.~Tagliasacchi, and V.~Sitzmann, ``pixelsplat: 3d gaussian splats from image pairs for scalable generalizable 3d reconstruction,'' in \emph{Proceedings of the IEEE/CVF Conference on Computer Vision and Pattern Recognition (CVPR)}, 2024, pp. 19\,457--19\,467.

\bibitem{mvsplat}
Y.~Chen, H.~Xu, C.~Zheng, B.~Zhuang, M.~Pollefeys, A.~Geiger, T.-J. Cham, and J.~Cai, ``Mvsplat: Efficient 3d gaussian splatting from sparse multi-view images,'' in \emph{European Conference on Computer Vision (ECCV)}, 2024, pp. 370--386.

\bibitem{drivingforward}
Q.~Tian, X.~Tan, Y.~Xie, and L.~Ma, ``Drivingforward: Feed-forward 3d gaussian splatting for driving scene reconstruction from flexible surround-view input,'' in \emph{Proceedings of the AAAI Conference on Artificial Intelligence (AAAI)}, vol.~39, no.~7, 2025, pp. 7374--7382.

\bibitem{vggt}
J.~Wang, M.~Chen, N.~Karaev, A.~Vedaldi, C.~Rupprecht, and D.~Novotny, ``Vggt: Visual geometry grounded transformer,'' in \emph{Proceedings of the IEEE/CVF Conference on Computer Vision and Pattern Recognition (CVPR)}, 2025, pp. 5294--5306.

\bibitem{anysplat}
L.~Jiang, Y.~Mao, L.~Xu, T.~Lu, K.~Ren, Y.~Jin, X.~Xu, M.~Yu, J.~Pang, F.~Zhao \emph{et~al.}, ``Anysplat: Feed-forward 3d gaussian splatting from unconstrained views,'' \emph{ACM Transactions on Graphics (TOG)}, vol.~44, no.~6, pp. 257:1--257:16, 2025.

\bibitem{semanticnerf}
S.~Zhi, T.~Laidlow, S.~Leutenegger, and A.~J. Davison, ``In-place scene labelling and understanding with implicit scene representation,'' in \emph{Proceedings of the IEEE/CVF International Conference on Computer Vision (ICCV)}, 2021, pp. 15\,838--15\,847.

\bibitem{dff}
S.~Kobayashi, E.~Matsumoto, and V.~Sitzmann, ``Decomposing nerf for editing via feature field distillation,'' \emph{Advances in Neural Information Processing Systems (NeurIPS)}, vol.~35, pp. 23\,311--23\,330, 2022.

\bibitem{iggt}
H.~Li, Z.~Zou, F.~Liu, X.~Zhang, F.~Hong, Y.~Cao, Y.~LAN, M.~Zhang, G.~YU, D.~Zhang, and Z.~Liu, ``Iggt: Instance-grounded geometry transformer for semantic 3d reconstruction,'' in \emph{International Conference on Learning Representations (ICLR)}, 2026.

\bibitem{ff3r}
C.~Zhou, R.~Wang, F.~Luo, M.~D. Pes{\'e}, Z.~Fan, Y.~Zhong, and S.~Huang, ``Ff3r: Feedforward feature 3d reconstruction from unconstrained views,'' \emph{arXiv preprint arXiv:2604.09862}, 2026.

\bibitem{unite}
S.~Koch, J.~Wald, H.~Matsuki, P.~Hermosilla, T.~Ropinski, and F.~Tombari, ``Unified semantic transformer for 3d scene understanding,'' \emph{Transactions on Machine Learning Research}, 2026.

\bibitem{fast3dis}
C.~Li, X.~Huang, S.-F. Chng, H.~Zhan, Q.~Yan, and Y.~Xu, ``Fast3dis: Feed-forward anchored scene transformer for 3d instance segmentation,'' \emph{arXiv preprint arXiv:2603.25993}, 2026.

\bibitem{semanticsplat}
Q.~Li, J.~Sun, L.~An, Z.~Su, H.~Zhang, and Y.~Liu, ``Semanticsplat: Feed-forward 3d scene understanding with language-aware gaussian fields,'' \emph{arXiv preprint arXiv:2506.09565}, 2025.

\bibitem{spatialsplat}
Y.~Sheng, J.~Deng, X.~Zhang, Y.~Zhang, B.~Hua, Y.~Zhang, and J.~Ji, ``Spatialsplat: Efficient semantic 3d from sparse unposed images,'' in \emph{Proceedings of the IEEE/CVF International Conference on Computer Vision (ICCV)}, 2025, pp. 26\,404--26\,414.

\bibitem{gsemsplat}
X.~Wang, C.~Lan, H.~Zhu, Z.~Chen, and Y.~Lu, ``Gsemsplat: Generalizable semantic 3d gaussian splatting from uncalibrated image pairs,'' \emph{arXiv preprint arXiv:2412.16932}, 2024.

\bibitem{uniforward}
Q.~Tian, X.~Tan, J.~Gong, Y.~Xie, and L.~Ma, ``Uniforward: Unified 3d scene and semantic field reconstruction via feed-forward gaussian splatting from only sparse-view images,'' \emph{arXiv preprint arXiv:2506.09378}, 2025.

\bibitem{fleg}
Q.~Tian, X.~Tan, J.~Ying, X.~Wang, Y.~Xie, and L.~Ma, ``Fleg: Feed-forward language embedded gaussian splatting from any views,'' \emph{arXiv preprint arXiv:2512.17541}, 2025.

\bibitem{siu3r}
Q.~Xu, D.~Wei, L.~Zhao, W.~Li, Z.~Huang, S.~Ji, and P.~Liu, ``Siu3r: Simultaneous scene understanding and 3d reconstruction beyond feature alignment,'' in \emph{Advances in Neural Information Processing Systems (NeurIPS)}, vol.~38, 2025, pp. 110\,800--110\,830.

\bibitem{clip}
A.~Radford, J.~W. Kim, C.~Hallacy, A.~Ramesh, G.~Goh, S.~Agarwal, G.~Sastry, A.~Askell, P.~Mishkin, J.~Clark \emph{et~al.}, ``Learning transferable visual models from natural language supervision,'' in \emph{International Conference on Machine Learning (ICML)}, 2021, pp. 8748--8763.

\bibitem{lseg}
B.~Li, K.~Q. Weinberger, S.~Belongie, V.~Koltun, and R.~Ranftl, ``Language-driven semantic segmentation,'' in \emph{International Conference on Learning Representations (ICLR)}, 2022.

\bibitem{scenesplat}
Y.~Li, Q.~Ma, R.~Yang, H.~Li, M.~Ma, B.~Ren, N.~Popovic, N.~Sebe, E.~Konukoglu, T.~Gevers \emph{et~al.}, ``Scenesplat: Gaussian splatting-based scene understanding with vision-language pretraining,'' in \emph{Proceedings of the IEEE/CVF International Conference on Computer Vision (ICCV)}, 2025, pp. 4961--4972.

\bibitem{scenesplat++}
M.~Ma, Q.~Ma, Y.~Li, J.~Cheng, R.~Yang, B.~Ren, N.~Popovic, M.~Wei, N.~Sebe, E.~Konukoglu \emph{et~al.}, ``Scenesplat++: A large dataset and comprehensive benchmark for language gaussian splatting,'' \emph{Advances in Neural Information Processing Systems (NeurIPS)}, vol.~38, 2025.

\bibitem{gpt4o}
\BIBentryALTinterwordspacing
{OpenAI}, ``{GPT-4o System Card},'' Aug. 2024. [Online]. Available: \url{https://openai.com/index/gpt-4o-system-card/}
\BIBentrySTDinterwordspacing

\bibitem{qwen3vl}
S.~Bai, Y.~Cai, R.~Chen, K.~Chen, X.~Chen, Z.~Cheng, L.~Deng, W.~Ding, C.~Gao, C.~Ge \emph{et~al.}, ``Qwen3-vl technical report,'' \emph{arXiv preprint arXiv:2511.21631}, 2025.

\bibitem{sa2va}
H.~Yuan, X.~Li, T.~Zhang, Z.~H. Huang, S.~Xu, S.~Ji, Y.~Tong, L.~Qi, J.~Feng, and M.-H. Yang, ``Sa2va: Marrying sam2 with llava for dense grounded understanding of images and videos,'' \emph{arXiv preprint}, 2025.

\bibitem{efficientmllm}
Y.~Jin, J.~Li, T.~Gu, Y.~Liu, B.~Zhao, J.~Lai, Z.~Gan, Y.~Wang, C.~Wang, X.~Tan \emph{et~al.}, ``Efficient multimodal large language models: A survey,'' \emph{Visual Intelligence}, vol.~3, no.~1, p.~27, 2025.

\bibitem{refersplat}
S.~He, G.~Jie, C.~Wang, Y.~Zhou, S.~Hu, G.~Li, and H.~Ding, ``Refersplat: Referring segmentation in 3d gaussian splatting,'' in \emph{International Conference on Machine Learning (ICML)}, 2025, pp. 22\,456--22\,467.

\bibitem{realm}
C.~Shi, M.~Chen, Y.~Mao, C.~Yang, X.~Hu, J.~Ding, and Z.~Yu, ``Realm: An mllm-agent framework for open world 3d reasoning segmentation and editing on gaussian splatting,'' in \emph{Proceedings of the IEEE/CVF Conference on Computer Vision and Pattern Recognition (CVPR)}, 2026, pp. 16\,779--16\,788.

\bibitem{dice}
F.~Milletari, N.~Navab, and S.-A. Ahmadi, ``V-net: Fully convolutional neural networks for volumetric medical image segmentation,'' in \emph{International Conference on 3D Vision (3DV)}, 2016, pp. 565--571.

\bibitem{focal}
T.-Y. Lin, P.~Goyal, R.~Girshick, K.~He, and P.~Doll{\'a}r, ``Focal loss for dense object detection,'' in \emph{Proceedings of the IEEE International Conference on Computer Vision (ICCV)}, 2017, pp. 2980--2988.

\bibitem{hdbscan}
L.~McInnes, J.~Healy, S.~Astels \emph{et~al.}, ``hdbscan: Hierarchical density based clustering.'' \emph{J. Open Source Softw.}, vol.~2, no.~11, p. 205, 2017.

\bibitem{sam2}
N.~Ravi, V.~Gabeur, Y.-T. Hu, R.~Hu, C.~Ryali, T.~Ma, H.~Khedr, R.~R{\"a}dle, C.~Rolland, L.~Gustafson \emph{et~al.}, ``Sam 2: Segment anything in images and videos,'' in \emph{International Conference on Learning Representations (ICLR)}, 2025.

\bibitem{gsplat}
V.~Ye, R.~Li, J.~Kerr, M.~Turkulainen, B.~Yi, Z.~Pan, O.~Seiskari, J.~Ye, J.~Hu, M.~Tancik, and A.~Kanazawa, ``gsplat: An open-source library for gaussian splatting,'' \emph{Journal of Machine Learning Research}, vol.~26, no.~34, pp. 1--17, 2025.

\bibitem{vllm}
W.~Kwon, Z.~Li, S.~Zhuang, Y.~Sheng, L.~Zheng, C.~H. Yu, J.~E. Gonzalez, H.~Zhang, and I.~Stoica, ``Efficient memory management for large language model serving with pagedattention,'' in \emph{Proceedings of the ACM Symposium on Operating Systems Principles (SOSP)}, 2023.

\bibitem{scannet}
A.~Dai, A.~X. Chang, M.~Savva, M.~Halber, T.~Funkhouser, and M.~Nie{\ss}ner, ``Scannet: Richly-annotated 3d reconstructions of indoor scenes,'' in \emph{Proceedings of the IEEE/CVF Conference on Computer Vision and Pattern Recognition (CVPR)}, 2017.

\bibitem{replica}
J.~Straub, T.~Whelan, L.~Ma, Y.~Chen, E.~Wijmans, S.~Green, J.~J. Engel, R.~Mur-Artal, C.~Ren, S.~Verma \emph{et~al.}, ``The replica dataset: A digital replica of indoor spaces,'' \emph{arXiv preprint arXiv:1906.05797}, 2019.

\bibitem{ssim}
Z.~Wang, A.~C. Bovik, H.~R. Sheikh, and E.~P. Simoncelli, ``Image quality assessment: from error visibility to structural similarity,'' \emph{IEEE Transactions on Image Processing (TIP)}, vol.~13, no.~4, pp. 600--612, 2004.

\bibitem{lpips}
R.~Zhang, P.~Isola, A.~A. Efros, E.~Shechtman, and O.~Wang, ``The unreasonable effectiveness of deep features as a perceptual metric,'' in \emph{Proceedings of the IEEE/CVF Conference on Computer Vision and Pattern Recognition (CVPR)}, 2018, pp. 586--595.

\bibitem{openseg}
G.~Ghiasi, X.~Gu, Y.~Cui, and T.-Y. Lin, ``Scaling open-vocabulary image segmentation with image-level labels,'' in \emph{European Conference on Computer Vision (ECCV)}, 2022, pp. 540--557.

\end{thebibliography}

\end{document}